\documentclass[letterpaper]{article} % DO NOT CHANGE THIS
\usepackage{aaai2027}  % DO NOT CHANGE THIS
\usepackage[hyphens]{url}  % DO NOT CHANGE THIS
\usepackage{graphicx} % DO NOT CHANGE THIS
\usepackage{natbib}  % DO NOT CHANGE THIS AND DO NOT ADD ANY OPTIONS TO IT
\usepackage{caption} % DO NOT CHANGE THIS AND DO NOT ADD ANY OPTIONS TO IT
\usepackage{amsmath,amsfonts,bm}

\def\eqref#1{equation~\ref{#1}}
\def\1{\bm{1}}

\def\vc{{\bm{c}}}

\def\ve{{\bm{e}}}

\def\vw{{\bm{w}}}
\def\vx{{\bm{x}}}

\def\mE{{\bm{E}}}

\def\mH{{\bm{H}}}
\def\mI{{\bm{I}}}

\def\mR{{\bm{R}}}

\def\mU{{\bm{U}}}

\def\mW{{\bm{W}}}
\def\mX{{\bm{X}}}
\def\mY{{\bm{Y}}}

\DeclareMathAlphabet{\mathsfit}{\encodingdefault}{\sfdefault}{m}{sl}
\SetMathAlphabet{\mathsfit}{bold}{\encodingdefault}{\sfdefault}{bx}{n}

\def\sC{{\mathbb{C}}}

\def\sF{{\mathbb{F}}}

\def\sI{{\mathbb{I}}}

\def\sQ{{\mathbb{Q}}}

\newcommand{\R}{\mathbb{R}}

\usepackage{booktabs}
\usepackage{multirow}
\usepackage{algorithm}
\usepackage{algorithmic}
\usepackage{subcaption}
\usepackage{amssymb}
\usepackage{tikz}
\usetikzlibrary{arrows.meta,positioning,calc,decorations.pathreplacing}
\usepackage[table]{xcolor}
\definecolor{soscalegray}{RGB}{235,235,245}
\newcommand{\soscalecell}[1]{\cellcolor{soscalegray}#1}
\graphicspath{{../SoScale/}}

\title{SoScale: Second-Order Scale Search for FP4 Quantization}
\author {
    Pengxiang Zhao,
    Xing Li,
    Hui-Ling Zhen,
    Weizhe Lin,
    Xianzhi Yu,
    Zhenhua Dong
}
\affiliations {
    Huawei Technologies Co., Ltd.\\
}

\begin{document}

\maketitle

\begin{abstract}
Low-bit quantization reduces the memory footprint and inference cost of large language models (LLMs), yet accurate FP4 quantization remains challenging. Practical formats such as MXFP4, NVFP4, and HiF4 pair a 4-bit floating-point payload with block-wise or hierarchical scale codes, making accuracy highly sensitive to how each local distribution is scaled. Default maximum-absolute-value scaling preserves dynamic range, but can spend scarce FP4 levels on rarely occupied outlier regions and leave dense tensor values poorly resolved. Existing scale-search methods minimize local element-wise reconstruction error, optimizing tensor fidelity rather than layer-output fidelity. We propose \textbf{SoScale}, a second-order weight scale-code search method for hierarchical FP4 quantization. SoScale searches hardware-valid scale codes under a Hessian-aware layer reconstruction objective and couples this search with GPTQ-style error recovery, selecting scales by their impact on layer outputs. Across Qwen3 models, SoScale consistently reduces downstream quantization degradation under MXFP4, NVFP4, and HiF4.
\end{abstract}

\section{Introduction}

Large language models (LLMs) have demonstrated strong capabilities across a wide range of natural language processing tasks~\citep{openai2026chatgpt, deepseekai2026deepseekv4, qwen3.5}, but their deployment remains constrained by substantial memory footprint and inference cost. Post-training quantization has therefore become an important technique for reducing model size and improving serving efficiency without expensive retraining~\citep{frantar2023gptq, xiao2023smoothquant, lin2024awq, shao2024omniquant, zhao2025ganq}. Among recent low-bit formats, FP4 is appealing for LLM inference because it compresses each value to four bits while retaining exponent-based dynamic range and enabling hardware acceleration for low-bit floating-point computation. Practical FP4 formats, such as MXFP4~\citep{rouhani2023microscaling}, NVFP4~\citep{alvarez2025nvfp4}, and HiF4~\citep{luo2026hifloat4formatlanguagemodel}, further introduce hierarchical scaling mechanisms to better accommodate the non-uniform distributions of LLM weights and activations.

\begin{figure*}[t]
    \centering
    \begin{subfigure}[t]{0.32\linewidth}
        \centering
        \includegraphics[width=\linewidth]{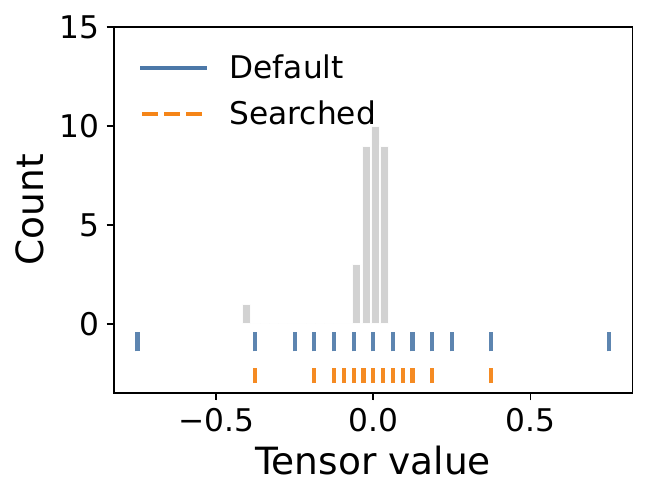}
        \caption{Scale-code alignment.}
        \label{fig:mxfp4-scale-grid}
    \end{subfigure}
    \hfill
    \begin{subfigure}[t]{0.32\linewidth}
        \centering
        \includegraphics[width=\linewidth]{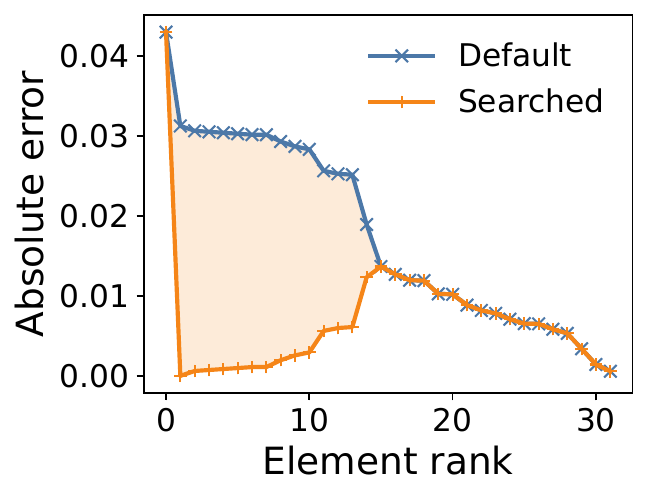}
        \caption{Element-wise error.}
        \label{fig:mxfp4-scale-error}
    \end{subfigure}
    \hfill
    \begin{subfigure}[t]{0.32\linewidth}
        \centering
        \includegraphics[width=\linewidth]{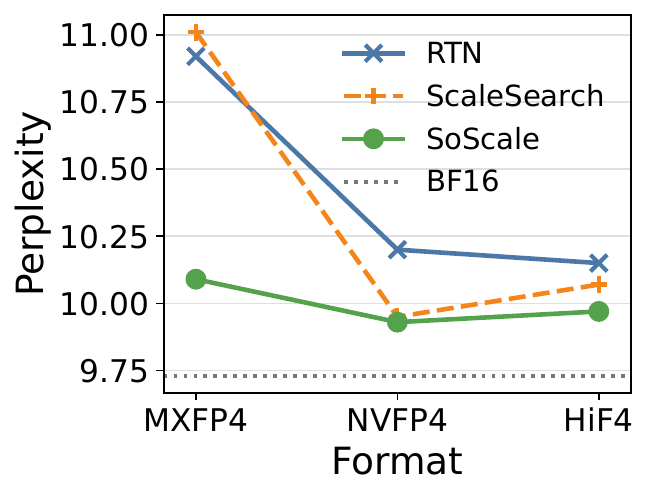}
        \caption{PPL comparison.}
        \label{fig:ppl-comparison}
    \end{subfigure}
    \caption{
    Scale-code alignment and model-level impact.
    On a real Qwen3-8B weight block, scale search changes the MXFP4 scale from $0.125$ to $0.0625$.
    (a) Default scaling preserves a wider range but leaves extreme levels unused, while the searched scale concentrates levels near dense tensor values.
    (b) The searched scale reduces element-wise reconstruction error for most elements.
    (c) Local element-wise scale search does not consistently reduce perplexity, whereas SoScale does.
    }
    \label{fig:mxfp4-scale-alignment}
\end{figure*}

Despite this promise, accurate quantization with hierarchical FP4 formats remains highly sensitive to scale-code selection. Since the FP4 payload provides only 16 representable values, the block scale determines where these quantization levels fall on the real-valued axis. When a block contains outliers, the standard maximum-absolute-value scaling rule expands the grid to cover the largest-magnitude element. This preserves range, but can spend levels on rarely occupied regions and reduce resolution near the dense mass of tensor values. Figure~\ref{fig:mxfp4-scale-grid} illustrates this effect on a real Qwen3-8B MXFP4 weight block. With default scaling, the grid is stretched by the outliers, yet the two outermost representable levels receive no assigned values because the outliers are rounded to the next-inner levels. This leaves fewer useful levels near zero, whereas the searched scale contracts the grid toward the occupied region and improves resolution for dense values. Figure~\ref{fig:mxfp4-scale-error} provides a complementary error view: after sorting values within the block, the absolute errors at the two tails remain nearly unchanged, while errors in the middle region, corresponding to values near zero, drop substantially under the searched scale.

The example above suggests that adapting the scale can be an effective way to reduce FP4 quantization error, and prior work has explored several related directions. Some methods reshape tensor distributions before quantization~\citep{xiao2023smoothquant, ashkboos2024quarot, liu2025spinquant, shao2025blockrotationneedmxfp4}; others compensate residual quantization errors after or during quantization~\citep{frantar2023gptq, meng2026arcquantboostingnvfp4quantization}; and FP4-specific methods adapt block scales with format-specific heuristics or local distortion objectives~\citep{cook2026sixaccuratenvfp4quantization, gupta2026searchblockfloatingpoint}. These approaches show that distribution-aware FP4 quantization is important, but most scale-selection objectives remain local: they choose a scale code by how well it reconstructs a block, even though the code ultimately matters through its effect on the layer output. As shown in Figure~\ref{fig:ppl-comparison}, reducing local element-wise error does not necessarily reduce model-level perplexity. This motivates treating scale selection as a layer reconstruction problem rather than a local tensor fitting problem.

In this work, we propose \textbf{SoScale}, a second-order post-training quantization (PTQ) method for weight and activation scale-code selection in hierarchical FP4 quantization. SoScale operates on a fixed pretrained model, without retraining or modifying the target FP4 format; instead, it searches within the format's legal scale-code space and chooses codes by minimizing a Hessian-weighted layer reconstruction loss. At quantization time, SoScale formulates scale selection as a sequential second-order reconstruction problem: after each block is committed, its induced error is propagated through the inverse-Hessian factor, so later blocks are optimized in a compensated parameter space rather than against the original weights~\citep{frantar2023gptq}. This candidate-set view makes the method format-agnostic: given the legal scale-code set and quantizer of a hierarchical FP4 format, the same Hessian-aware search and recovery procedure can be applied without format-specific redesign. Since weight quantization is performed offline and the resulting representation preserves the target FP4 encoding, SoScale does not introduce serving-time search and remains compatible with hardware inference paths that already support the corresponding format. We instantiate this general formulation for MXFP4, NVFP4, and HiF4, where it remains complementary to preprocessing methods such as block Hadamard rotations~\citep{shao2025blockrotationneedmxfp4}.

We evaluate SoScale on Qwen3-8B and Qwen3-14B~\citep{yang2025qwen3technicalreport} across three hierarchical FP4 formats: MXFP4, NVFP4, and HiF4. Across downstream benchmarks, the best SoScale variants achieve the highest average score in every model-format group, outperforming RTN, GPTQ~\citep{frantar2023gptq}, local ScaleSearch~\citep{gupta2026searchblockfloatingpoint}, and BlockRotation~\citep{ashkboos2024quarot} baselines. On Qwen3-14B, these variants retain 97.17\% of BF16 average accuracy under MXFP4, 99.20\% under NVFP4, and 99.40\% under HiF4. These results indicate that optimizing scale codes with second-order layer information recovers accuracy that local tensor-level objectives miss.

Our contributions are summarized as follows:
\begin{itemize}
    \item We identify an outlier-induced scale mismatch in hierarchical FP4 quantization: default range-preserving scales can reserve scarce FP4 levels for regions that receive no assignments, reducing resolution for dense weight values.
    \item We introduce \textbf{SoScale}, an offline scale-code optimizer that preserves the target FP4 format, combines Hessian-weighted layer reconstruction with quantization error compensation, and retains compatibility with existing FP4 hardware inference paths.
    \item We show consistent gains over representative baselines across two Qwen3 models and three FP4 formats.
\end{itemize}

\section{Related Work}

\paragraph{Hierarchical FP4 formats.}
Low-bit floating-point formats reduce LLM inference cost while preserving exponent-based dynamic range and enabling hardware acceleration. Practical FP4 formats, including MXFP4~\citep{rouhani2023microscaling}, NVFP4~\citep{alvarez2025nvfp4}, and HiF4~\citep{luo2026hifloat4formatlanguagemodel}, pair a 4-bit payload with block-wise or hierarchical scales; their main structural differences are summarized in Table~\ref{tab:format-comparison}. Despite differences in scale hierarchy, granularity, and codebooks, these formats share the same abstraction: the scale code places the FP4 payload levels on the real-valued axis for each block. Consequently, scale selection becomes a primary source of accuracy loss when block distributions are heavy-tailed or contain outliers.

\begin{table}[t]
\centering
\small
\setlength{\tabcolsep}{4pt}
\begin{tabular}{lccc}
\toprule
 & \textbf{MXFP4} & \textbf{NVFP4} & \textbf{HiF4} \\
\midrule
Element format & E2M1 & E2M1 & E1M2 \\
Scale hierarchy & E8M0 & FP32 + E4M3 & E6M2 + E1 + E1 \\
Granularity & 32 & 16 + tensor & 64 / 8 / 4 \\
Bits / element & 4.25 & $\gtrsim$4.50 & 4.50 \\
\bottomrule
\end{tabular}
\caption{
Compact comparison of hierarchical FP4 formats considered in this work.
}
\label{tab:format-comparison}
\end{table}

\paragraph{Post-training quantization methods.}
PTQ reduces the memory and compute cost of pretrained LLMs without expensive retraining. Simple round-to-nearest (RTN) quantization is widely used as an efficient baseline, but often degrades accuracy at very low bit-widths. Calibration-based PTQ methods improve low-bit accuracy by using activation statistics, saliency estimates, or second-order approximations to preserve layer behavior~\citep{frantar2023gptq, lin2024awq, shao2024omniquant, zhao2025ganq}. A complementary line smooths, rotates, or rebalances weights and activations before quantization to reduce outlier-induced difficulty~\citep{xiao2023smoothquant, ashkboos2024quarot, liu2025spinquant}. A common lesson is that quantization errors should be evaluated by how they affect layer outputs, not only by element-wise tensor distortion. SoScale adopts this layer-centric view, but applies it to the discrete scale codes of hierarchical FP4 formats.

\paragraph{FP4 methods.}
Recent FP4 methods typically handle outlier sensitivity through preprocessing, compensation, or scale adaptation. Block Hadamard rotations shrink each MXFP4 block before quantization, but rely on a downstream scale rule~\citep{shao2025blockrotationneedmxfp4}. ARCQuant repairs NVFP4 errors with reordering and residual channels, adding compensation instead of optimizing scales~\citep{meng2026arcquantboostingnvfp4quantization}. The closest work adapts scales directly: Four Over Six switches between two NVFP4 rules~\citep{cook2026sixaccuratenvfp4quantization}, and ScaleSearch minimizes local mean squared error (MSE) over block scales~\citep{gupta2026searchblockfloatingpoint}. These methods establish scale selection as important, but remain tied to format-specific rules or local tensor loss. SoScale uses a Hessian-weighted layer reconstruction loss to search legal scale codes across MXFP4, NVFP4, and HiF4, with GPTQ recovery.

\section{Methodology}
This section presents \textbf{SoScale}, a weight scale-code search method for hierarchical FP4 quantization. For each linear layer, SoScale selects legal scale codes with a Hessian-weighted layer reconstruction objective and formulates blockwise quantization as a sequential second-order reconstruction process with error recovery. We first formulate scale-code search, then describe error recovery, and finally summarize the full pipeline.

\subsection{Hessian-Aware Scale-Code Search}

Modern LLMs are dominated by linear projections in attention and feed-forward modules, making layer reconstruction a natural objective for post-training quantization. Consider a linear layer with calibration inputs $\mX \in \R^{T \times n}$ and full-precision weights $\mW \in \R^{m \times n}$:
\begin{equation}
\mY = \mX \mW^\top ,
\end{equation}
where $T$ denotes the number of calibration tokens, and $n$ and $m$ denote the input and output dimensions. Given quantized weights $\widehat{\mW}$, SoScale measures quantization quality by the calibration-based layer reconstruction error
\begin{equation}
\begin{aligned}
\mathcal{L}(\mW,\widehat{\mW})
&=
\left\|
\mX\mW^\top - \mX\widehat{\mW}^\top
\right\|_F^2 \\
&=
\sum_{i=1}^{m}
\ve_i^\top
\mH
\ve_i,
\end{aligned}
\end{equation}
where $\ve_i=\vw_i-\widehat{\vw}_i$, $\vw_i$ and $\widehat{\vw}_i$ are the $i$-th rows of $\mW$ and $\widehat{\mW}$, and $\mH=\mX^\top\mX$ is the empirical Hessian. Unlike element-wise distortion metrics, this objective does not treat all weight errors equally: errors that induce larger changes in the calibrated layer outputs receive larger penalties.

The reconstruction objective specifies the desired layer behavior, while the target FP4 format defines the discrete feasible set for $\widehat{\mW}$. In hierarchical FP4 quantization, each row of $\mW$ is partitioned into blocks along the input dimension, and each block is quantized by combining a low-bit FP4 payload with one or more scale codes, as summarized in Table~\ref{tab:format-comparison}. Let $\vw_{i,b} \in \R^{B}$ denote block $b$ of row $i$, where $B$ is the format-specific block size. For a target format $\mathcal{F}$, we write the quantized block as
\begin{equation}
\widehat{\vw}_{i,b}(\vc)
=
Q_{\mathcal{F}}(\vw_{i,b}; \vc),
\end{equation}
where $Q_{\mathcal{F}}$ is the format-specific FP4 quantizer and $\vc \in \sC_{\mathcal{F}}$ denotes a legal scale configuration. For each candidate $\vc$, the payload values are obtained by the standard rounding and clipping rule of the target format; SoScale only changes which legal scale codes are used. For concreteness, consider MXFP4. Each block uses an E2M1 payload and an E8M0 block-scale exponent. Let $c \in \sC_{\mathrm{MX}}$ denote the exponent code and let $s(c)=2^c$ be the corresponding block scale. The quantized block is
\begin{equation}
\widehat{\vw}_{i,b}(c)
=
s(c)\,
\Pi_{\mathrm{E2M1}}
\left(
\frac{\vw_{i,b}}{s(c)}
\right),
\end{equation}
where $\Pi_{\mathrm{E2M1}}(\cdot)$ denotes rounding and clipping to the E2M1 payload codebook. For a fixed E2M1 codebook, the exponent code $c$ induces the scaled representable set $s(c)\mathcal{C}_{\mathrm{E2M1}}$: increasing $s(c)$ expands the covered range but coarsens the spacing between levels, while decreasing $s(c)$ contracts the range and improves resolution for smaller magnitudes, as illustrated in Figure~\ref{fig:mxfp4-scale-grid}. The same notation covers the other hierarchical formats considered in this work: for NVFP4, SoScale searches the E4M3 block-scale code while keeping the FP32 tensor scale fixed; for HiF4, it searches the E6M2 scale code and derives the E1 scales accordingly.

After fixing the FP4 payload and legal scale space, scale-code optimization reduces to selecting one code for each block. The ideal choice would minimize the layer reconstruction loss above over all scale codes, but the full Hessian couples errors across blocks and makes joint search combinatorial. SoScale therefore uses a block-local second-order surrogate aligned with the FP4 layout. Let $\sI_b$ denote the input indices of block $b$, and let
\begin{equation}
\mH_b = \mH_{\sI_b,\sI_b}
\end{equation}
be the corresponding Hessian submatrix. For a candidate scale configuration $\vc$, define the row-block error
\begin{equation}
\ve_{i,b}(\vc)
=
\vw_{i,b} - \widehat{\vw}_{i,b}(\vc).
\end{equation}
SoScale scores $\vc$ by
\begin{equation}
\ell_{i,b}(\vc)
=
\ve_{i,b}(\vc)^\top
\mH_b
\ve_{i,b}(\vc),
\end{equation}
The matrix $\mH_b$ is what distinguishes this score from local MSE: if $\mH_b$ were the identity, $\ell_{i,b}$ would reduce to squared block error, whereas the empirical Hessian weights errors and correlations according to the calibration activations. For each row-block pair $(i,b)$, SoScale selects
\begin{equation}
\vc_{i,b}^{\star}
=
\arg\min_{\vc \in \sC_{i,b}}
\ell_{i,b}(\vc),
\end{equation}
where $\sC_{i,b}$ is formed by applying a small format-specific offset window around the default scale code and clipping invalid codes to the legal range of the target format.

\subsection{Second-Order Error Recovery}

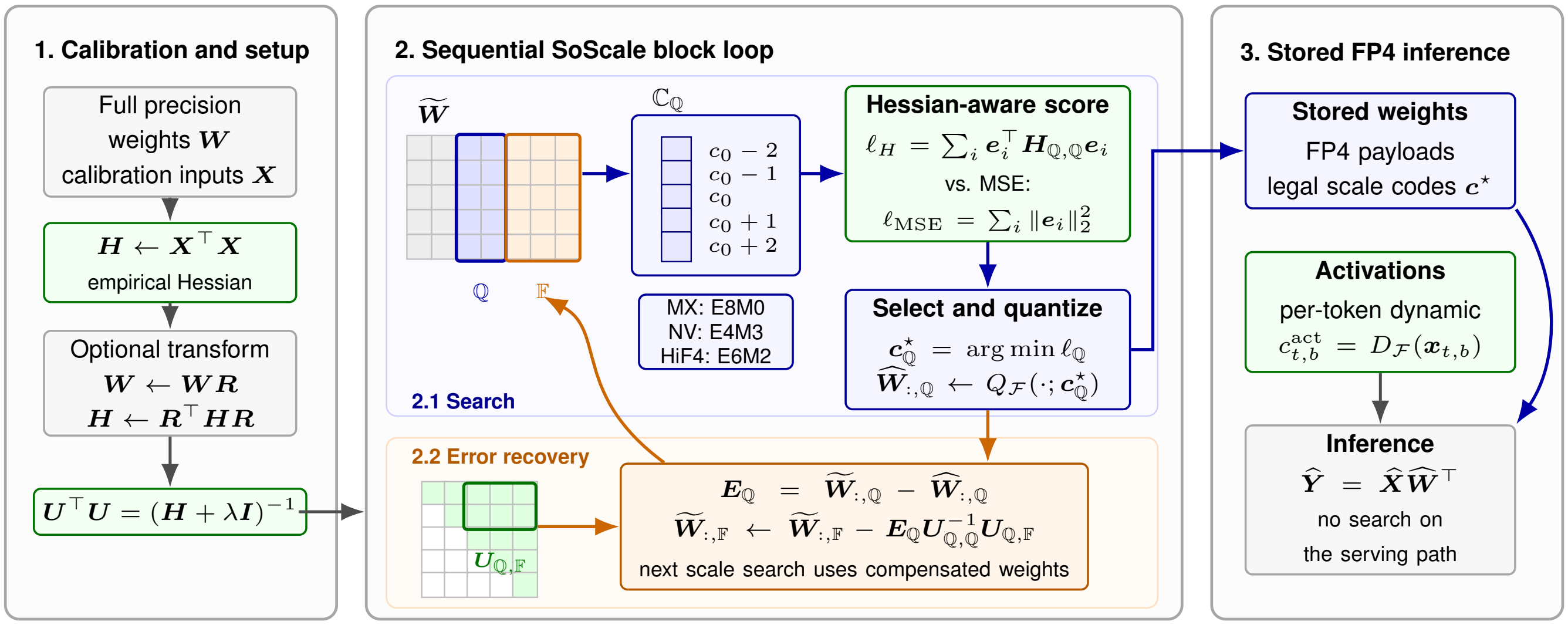
\begin{figure*}[t]
\centering
\resizebox{\linewidth}{!}{
\begin{tikzpicture}[
    font=\sffamily\footnotesize,
    >=Latex,
    stage/.style={draw=black!35, rounded corners=4pt, fill=black!1, line width=0.8pt},
    title/.style={font=\sffamily\bfseries\scriptsize, anchor=west},
    box/.style={draw=black!35, rounded corners=2pt, fill=white, align=center, inner sep=2.5pt, line width=0.6pt, font=\sffamily\scriptsize},
    bluebox/.style={box, draw=blue!60!black, fill=blue!4},
    greenbox/.style={box, draw=green!45!black, fill=green!5},
    orangebox/.style={box, draw=orange!70!black, fill=orange!8},
    graybox/.style={box, fill=black!3},
    phasepanel/.style={draw=black!15, rounded corners=3pt, fill=black!1, line width=0.45pt},
    cell/.style={draw=black!25, minimum width=0.25cm, minimum height=0.25cm, inner sep=0pt},
    wcell/.style={cell, fill=black!7},
    qcell/.style={cell, fill=blue!12},
    fcell/.style={cell, fill=orange!12},
    hcell/.style={cell, fill=green!12},
    zcell/.style={cell, fill=white},
    codecell/.style={draw=blue!45!black, fill=blue!9, minimum width=0.30cm, minimum height=0.30cm, inner sep=0pt},
    arrow/.style={->, line width=0.85pt, draw=black!70},
    bluearr/.style={->, line width=0.9pt, draw=blue!65!black},
    orangearr/.style={->, line width=0.9pt, draw=orange!80!black}
]

% Stage containers.
\node[stage, anchor=south west, minimum width=3.35cm, minimum height=6.20cm] (prep) at (0,0.60) {};
\node[stage, anchor=south west, minimum width=8.25cm, minimum height=6.20cm] (loop) at (3.65,0.60) {};
\node[stage, anchor=south west, minimum width=3.55cm, minimum height=6.20cm] (out) at (12.20,0.60) {};
\node[phasepanel, anchor=south west, minimum width=7.80cm, minimum height=3.44cm, fill=blue!2, draw=blue!18] (searchphase) at (3.86,2.66) {};
\node[phasepanel, anchor=south west, minimum width=7.80cm, minimum height=1.72cm, fill=orange!3, draw=orange!25] (recoveryphase) at (3.86,0.72) {};
\node[anchor=west, font=\sffamily\bfseries\tiny, blue!55!black] at (4.00,2.82) {2.1 Search};
\node[anchor=west, font=\sffamily\bfseries\tiny, orange!70!black] at (4.00,2.24) {2.2 Error recovery};
\node[title] at (0.18,6.35) {1. Calibration and setup};
\node[title] at (3.83,6.35) {2. Sequential SoScale block loop};
\node[title] at (12.38,6.35) {3. Stored FP4 inference};

% Calibration and second-order preparation.
\node[graybox, minimum width=2.55cm] (inputs) at (1.68,5.44) {Full precision\\weights $\mW$\\calibration inputs $\mX$};
\node[greenbox, minimum width=2.55cm] (hessian) at (1.68,4.22) {$\mH \leftarrow \mX^\top\mX$\\{\tiny empirical Hessian}};
\node[graybox, minimum width=2.55cm] (rotate) at (1.68,3.00) {Optional transform\\$\mW\leftarrow\mW\mR$\\$\mH\leftarrow\mR^\top\mH\mR$};
\node[greenbox, minimum width=2.55cm] (factor) at (1.68,1.70) {$\mU^\top\mU=(\mH+\lambda\mI)^{-1}$};
\draw[arrow] (inputs) -- (hessian);
\draw[arrow] (hessian) -- (rotate);
\draw[arrow] (rotate) -- (factor);

% Working weight matrix with current and future blocks.
\node[anchor=west, font=\sffamily\scriptsize] at (4.05,5.76) {$\widetilde{\mW}$};
\foreach \r in {0,...,4}{
  \foreach \c in {0,...,6}{
    \ifnum\c<2
      \node[wcell] (W-\r-\c) at ({4.20+0.25*\c},{5.38-0.25*\r}) {};
    \else
      \ifnum\c<4
        \node[qcell] (W-\r-\c) at ({4.20+0.25*\c},{5.38-0.25*\r}) {};
      \else
        \node[fcell] (W-\r-\c) at ({4.20+0.25*\c},{5.38-0.25*\r}) {};
      \fi
    \fi
  }
}
\draw[draw=blue!65!black, line width=0.95pt, rounded corners=1pt] (4.575,5.515) rectangle (5.075,4.235);
\draw[draw=orange!80!black, line width=0.85pt, rounded corners=1pt] (5.075,5.515) rectangle (5.825,4.235);
\node[anchor=north, font=\sffamily\tiny, blue!65!black] at (4.825,4.13) {$\sQ$};
\node[anchor=north, font=\sffamily\tiny, orange!80!black] at (5.45,4.13) {$\sF$};

% Candidate scale-code construction.
\node[anchor=west, font=\sffamily\scriptsize] at (6.43,5.86) {$\sC_{\sQ}$};
\node[bluebox, minimum width=1.70cm, minimum height=1.62cm] (codes) at (7.20,4.90) {};
\foreach \r/\code in {0/$c_0-2$,1/$c_0-1$,2/$c_0$,3/$c_0+1$,4/$c_0+2$}{
  \node[codecell] at (6.80,{5.34-0.24*\r}) {};
  \node[anchor=west, font=\sffamily\tiny] at (7.00,{5.34-0.24*\r}) {\code};
}
\node[bluebox, text width=1.45cm, inner sep=1.5pt, font=\sffamily\tiny] (formats) at (7.20,3.52) {MX: E8M0\\NV: E4M3\\HiF4: E6M2};

% Hessian-aware scoring and quantization.
\node[greenbox, text width=2.70cm] (score) at (9.95,5.22) {
\textbf{Hessian-aware score}\\[0.5mm]
$\ell_H=\sum_i \ve_i^\top\mH_{\sQ,\sQ}\ve_i$\\[0.4mm]
{\tiny vs. MSE: $\ell_{\mathrm{MSE}}=\sum_i\|\ve_i\|_2^2$}
};
\node[bluebox, text width=2.70cm] (select) at (9.95,3.34) {
\textbf{Select and quantize}\\[0.5mm]
$\vc_{\sQ}^{\star}=\arg\min \ell_{\sQ}$\\
$\widehat{\mW}_{:,\sQ}\leftarrow Q_{\mathcal F}(\cdot;\vc_{\sQ}^{\star})$
};
\draw[bluearr] (5.85,5.12) -- ($(codes.west)+(0,0.22)$);
\draw[bluearr] ($(codes.east)+(0,0.22)$) -- ($(score.west)+(0,-0.10)$);
\draw[bluearr] (score) -- (select);

% GPTQ propagation subroutine.
% \node[anchor=west, font=\sffamily\scriptsize] at (4.05,2.60) {Second-order error propagation};
\node[anchor=west, font=\sffamily\tiny] at (4.25,2.22) {};
\foreach \r in {0,...,4}{
  \foreach \c in {0,...,4}{
    \ifnum\c<\r
      \node[zcell] (U-\r-\c) at ({4.35+0.23*\c},{1.88-0.23*\r}) {};
    \else
      \node[hcell] (U-\r-\c) at ({4.35+0.23*\c},{1.88-0.23*\r}) {};
    \fi
  }
}
\draw[draw=green!45!black, line width=0.85pt, rounded corners=1pt] (4.65,1.99) rectangle (5.38,1.52);
\node[anchor=north, font=\sffamily\tiny, green!45!black] at (5.02,1.41) {$\mU_{\sQ,\sF}$};
\node[orangebox, text width=4.55cm] (prop) at (8.60,1.55) {
$\mE_{\sQ}=\widetilde{\mW}_{:,\sQ}-\widehat{\mW}_{:,\sQ}$\\[0.5mm]
$\widetilde{\mW}_{:,\sF}\leftarrow
\widetilde{\mW}_{:,\sF}-\mE_{\sQ}\mU_{\sQ,\sQ}^{-1}\mU_{\sQ,\sF}$\\[0.5mm]
{\tiny next scale search uses compensated weights}
};
\draw[orangearr] (select.south) -- ($(prop.north)+(1.35,0)$);
\draw[orangearr] (5.40,1.55) -- (prop.west);
\draw[orangearr] ($(prop.north west)+(0.45,0)$) .. controls +(-0.95,0.70) and +(0.65,-0.35) .. (5.45,3.88);

% Stored quantized representation and inference-time path.
\node[bluebox, text width=2.55cm] (stored) at (13.92,5.35) {
\textbf{Stored weights}\\[0.5mm]
FP4 payloads\\
legal scale codes $\vc^\star$
};
\node[greenbox, text width=2.55cm] (acts) at (13.92,3.72) {
\textbf{Activations}\\[0.5mm]
per-token dynamic\\
$c^{\mathrm{act}}_{t,b}=D_{\mathcal F}(\vx_{t,b})$
};
\node[graybox, text width=2.55cm] (infer) at (13.92,1.82) {
\textbf{Inference}\\[0.5mm]
$\widehat{\mY}=\widehat{\mX}\widehat{\mW}^{\top}$\\
{\tiny no search on the serving path}
};
\draw[bluearr] (stored.south east) .. controls +(0.45,-0.65) and +(0.45,0.65) .. (infer.north east);
\draw[arrow] (acts) -- (infer);
\draw[bluearr] (select.east) -- ++(0.22,0) |- node[pos=0.25, above, font=\sffamily\tiny] {} (stored.west);
\draw[arrow] (2.98,1.70) -- (3.65,1.70);

\end{tikzpicture}
}
\caption{
Workflow of SoScale for one linear layer.
Calibration builds the empirical Hessian and damped inverse-Hessian factor.
In 2.1, SoScale searches legal FP4 scale codes with a Hessian-aware layer loss, whereas scale-only baselines such as ScaleSearch and Four Over Six use local MSE-style objectives without Hessian information; in 2.2, it propagates the selected block's error to future blocks.
Inference uses only stored FP4 payloads, legal scale codes, and lightweight per-token activation scales.
}
\label{fig:dynamic-scale-search-gptq}
\end{figure*}

The previous subsection defines how to choose scale codes for a fixed block of weights. In sequential quantization, however, committing one block introduces an error that can be partially absorbed by the remaining unquantized blocks. If later blocks were searched on the original weights, their scale choices would ignore this compensation. This coupling is natural for hierarchical FP4 formats because both scale selection and error recovery operate at the block level. SoScale therefore follows the GPTQ principle~\citep{frantar2023gptq}: after each block is quantized, it propagates a second-order correction to future blocks, and the next scale-code search is performed on the compensated weights.

Let $\widetilde{\mW}$ denote the working copy of the weights, initialized as $\widetilde{\mW}=\mW$. Following GPTQ, we compute an upper-triangular factor $\mU$ of the damped inverse Hessian,
\begin{equation}
\mU^\top \mU = \left(\mH + \lambda \mI\right)^{-1},
\end{equation}
where $\lambda$ is a damping coefficient. At one sequential step, let $\sQ \subset \{1,\ldots,n\}$ denote the input-column indices of the current FP4 block, and let $\sF$ denote the columns belonging to the remaining future blocks. The slice $\widetilde{\mW}_{:,\sQ}$ contains the current block for all output rows. Each row in the slice has its own scale code or scale configuration, and we collect these row-wise codes in $\vc_{\sQ}$. Before committing the block, SoScale searches these codes on the compensated weights:
\begin{equation}
\vc_{\sQ}^{\star}
=
\arg\min_{\vc_{\sQ} \in \sC_{\sQ}}
\ell_{\sQ}(\vc_{\sQ}).
\end{equation}
Here $\ell_{\sQ}$ sums the row-block scores from the previous subsection over all output rows. Using the selected scale codes, SoScale commits the quantized block
\begin{equation}
\widehat{\mW}_{:,\sQ}
=
Q_{\mathcal{F}}
\left(
\widetilde{\mW}_{:,\sQ}; \vc_{\sQ}^{\star}
\right),
\end{equation}
and obtains the block error
\begin{equation}
\mE_{\sQ}
=
\widetilde{\mW}_{:,\sQ}
-
\widehat{\mW}_{:,\sQ}.
\end{equation}
The block error is then propagated to the future blocks through the back-substitution:
\begin{equation}
\widetilde{\mW}_{:,\sF}
\leftarrow
\widetilde{\mW}_{:,\sF}
-
\mE_{\sQ}
\mU_{\sQ,\sQ}^{-1}
\mU_{\sQ,\sF}.
\end{equation}
Here $\mU_{\sQ,\sQ}$ and $\mU_{\sQ,\sF}$ are the submatrices of $\mU$ indexed by the current block and the future input dimensions. This update changes only the working copy $\widetilde{\mW}$ used for subsequent blocks; the committed block $\widehat{\mW}_{:,\sQ}$ remains fixed. As a result, scale-code search and error recovery are interleaved: each block selects scales on weights that already include second-order compensation from earlier quantization errors.

\subsection{Dynamic Per-Token Activation Quantization}

SoScale optimizes the scale codes of weight blocks. Activations are handled by an online dynamic per-token FP4 rule under the target format. For token $t$, let $\vx_t$ denote its activation vector and let $\vx_{t,b}$ denote block $b$ under the target format. Each activation block is quantized as
\begin{equation}
\widehat{\vx}_{t,b}
=
Q_{\mathcal{F}}^{\mathrm{act}}
\left(
\vx_{t,b}; c^{\mathrm{act}}_{t,b}
\right),
\qquad
c^{\mathrm{act}}_{t,b}
=
D_{\mathcal{F}}(\vx_{t,b}),
\end{equation}
where $D_{\mathcal{F}}$ is the format-specific dynamic scale rule for activations. Because activation scales are computed online for every token, $D_{\mathcal{F}}$ is kept lightweight and selects $c^{\mathrm{act}}_{t,b}$ for each token block using an element-wise MSE criterion.

\subsection{Overall SoScale Framework}

\begin{algorithm}[t]
\caption{SoScale for one linear layer}
\label{alg:soscale}
\begin{algorithmic}[1]
\REQUIRE Full-precision weights $\mW$, calibration inputs $\mX$, target FP4 format $\mathcal{F}$, damping $\lambda$, optional transform $\mR$
\ENSURE Quantized weights $\widehat{\mW}$ and scale-code setting $\vc^\star$
\STATE Compute empirical Hessian $\mH \leftarrow \mX^\top \mX$
\IF{preprocessing transform $\mR$ is used}
    \STATE Transform weights and Hessian: $\mW \leftarrow \mW \mR$, $\mH \leftarrow \mR^\top \mH \mR$
\ENDIF
\STATE Initialize working weights $\widetilde{\mW} \leftarrow \mW$, quantized weights $\widehat{\mW}$, and scale codes $\vc^\star$
\STATE Compute upper-triangular factor $\mU^\top \mU = (\mH + \lambda\mI)^{-1}$
\FOR{each FP4 block $\sQ$ and future-block index set $\sF$}
    \STATE Construct legal candidate set $\sC_{\sQ}$
    \STATE Search on weights:
    $\vc_{\sQ}^{\star} \leftarrow \arg\min_{\vc_{\sQ} \in \sC_{\sQ}} \ell_{\sQ}(\vc_{\sQ})$
    \STATE Commit quantized block:
    $\widehat{\mW}_{:,\sQ} \leftarrow Q_{\mathcal{F}}(\widetilde{\mW}_{:,\sQ}; \vc_{\sQ}^{\star})$
    \STATE Compute block error $\mE_{\sQ} \leftarrow \widetilde{\mW}_{:,\sQ} - \widehat{\mW}_{:,\sQ}$
    \STATE Propagate error to future blocks:
    $\widetilde{\mW}_{:,\sF} \leftarrow \widetilde{\mW}_{:,\sF} - \mE_{\sQ}\mU_{\sQ,\sQ}^{-1}\mU_{\sQ,\sF}$
    \STATE Store $\vc_{\sQ}^{\star}$ in $\vc^\star$
\ENDFOR
\STATE \textbf{return} $\widehat{\mW}$ and $\vc^\star$
\end{algorithmic}
\end{algorithm}

Algorithm~\ref{alg:soscale} summarizes SoScale for one linear layer, and Figure~\ref{fig:dynamic-scale-search-gptq} visualizes the same workflow by separating Hessian-aware search (2.1) from error recovery (2.2). The algorithm builds the empirical Hessian from calibration activations, optionally applies a preprocessing transform to both weights and Hessian, and then processes FP4 input blocks sequentially. For each block, scale codes are searched on the current compensated weights, the selected quantized block is committed, and its error is propagated to future blocks through the back-substitution. Scale-only baselines such as ScaleSearch~\citep{gupta2026searchblockfloatingpoint} and Four Over Six~\citep{cook2026sixaccuratenvfp4quantization} mainly correspond to the 2.1 search stage and optimize local MSE-style objectives without Hessian information; SoScale instead scores legal scale codes with the Hessian-aware layer loss and then applies 2.2 recovery to update future blocks. The output is a target-format FP4 representation consisting of quantized payloads and legal scale codes. This weight quantization procedure is performed offline during calibration; inference uses only the stored FP4 payloads and scale codes.

\section{Experiments}

We evaluate SoScale on Qwen3-family LLMs under W4A4 quantization with three hierarchical FP4 formats: MXFP4, NVFP4, and HiF4. We report downstream benchmark accuracy against representative baselines.

\subsection{Experimental Setup}

\paragraph{Models.}
We evaluate SoScale on Qwen3-8B and Qwen3-14B~\citep{yang2025qwen3technicalreport}.

\paragraph{Quantization formats.}
We evaluate MXFP4, NVFP4, and HiF4, the three hierarchical FP4 formats summarized in Table~\ref{tab:format-comparison}. Our main setting is W4A4: weights are quantized by the compared weight quantizer, and activations use the online per-token FP4 rule described in the methodology. We quantize linear layers in transformer blocks, while keeping token embeddings, the LM head, and normalization layers in BF16. We also report a direct-activation setting for scale-search methods, where activations use direct RTN quantization; this isolates the weight-side contribution while keeping the activation path closer to hardware-native FP4 execution.

\paragraph{Baselines.}
We use \textbf{BF16} as the full-precision reference. \textbf{RTN} denotes the default format-specific round-to-nearest FP4 quantizer. \textbf{GPTQ} denotes GPTQ-style second-order weight quantization~\citep{frantar2023gptq}. Since GPTQ is originally a weight-only PTQ method, in the main W4A4 comparison we pair its quantized weights with the online per-token activation rule described above. \textbf{ScaleSearch} denotes local block-scale search that minimizes element-wise MSE~\citep{gupta2026searchblockfloatingpoint}. \textbf{BlockRotation} denotes the standalone block Hadamard preprocessing method proposed for MXFP4~\citep{shao2025blockrotationneedmxfp4}; in our experiments, we apply the same preprocessing to MXFP4, NVFP4, and HiF4. In tables, the suffix \textbf{w/ Had.} denotes applying block Hadamard preprocessing before the corresponding quantization method, following QuaRot~\citep{ashkboos2024quarot} and block rotation~\citep{shao2025blockrotationneedmxfp4}. We use these names consistently in the result tables.

\paragraph{Calibration and implementation.}
For methods that require calibration, we sample 128 sequences of 2,048 tokens from the first WikiText shard~\citep{merity2016pointer}, following a widely used PTQ setup~\citep{frantar2023gptq}. For SoScale, these calibration activations are used to estimate the empirical Hessian for the Hessian-aware scale-code search. We load Qwen3 checkpoints and tokenizers from their public Hugging Face repositories, use \texttt{datasets}~\citep{lhoest2021datasets} and \texttt{transformers}~\citep{wolf2020transformers} for data preprocessing and model execution, and implement all quantization routines in PyTorch~\citep{paszke2019pytorch}.

\paragraph{Evaluation benchmarks.}
We evaluate quantized models using the Language Model Evaluation Harness~\citep{gao2024lmevalharness} with batch size 64 on GSM8K~\citep{cobbe2021training}, HellaSwag~\citep{zellers2019hellaswag}, MMLU~\citep{hendrycks2021measuring}, ARC-Challenge and ARC-Easy~\citep{clark2018think}, BoolQ~\citep{clark2019boolq}, and PIQA~\citep{bisk2020piqa}. We report exact-match accuracy for GSM8K and accuracy for the remaining tasks. Unless otherwise noted, GSM8K is evaluated in a 5-shot setting and the other benchmarks are evaluated in a zero-shot setting. We report the average score across tasks and its relative ratio to the BF16 reference.

\subsection{Main Results}

\begin{table*}[!t]
\centering
\small
\renewcommand{\arraystretch}{0.89}
\setlength{\tabcolsep}{3pt}
\resizebox{\linewidth}{!}{
\begin{tabular}{l c l l ccccccc cc}
\toprule
\textbf{Model} & \textbf{Bit} & \textbf{Format} & \textbf{Method}
& \textbf{GSM8K} & \textbf{HellaSwag} & \textbf{MMLU}
& \textbf{ARC-C} & \textbf{ARC-E} & \textbf{BoolQ} & \textbf{PIQA}
& \textbf{Avg.} & \textbf{Rel.} \\
\midrule
\multirow{19}{*}{\raisebox{-0.45\normalbaselineskip}{Qwen3-8B}}
& -- & BF16 & --
& 0.8810 & 0.5713 & 0.7292 & 0.5538 & 0.8371 & 0.8663 & 0.7628 & 0.7431 & 1.0000 \\
\cmidrule(lr){2-13}
& \multirow{18}{*}{W4A4}
& \multirow{6}{*}{MXFP4}
& RTN & 0.8211 & 0.5292 & 0.6718 & 0.4923 & 0.7735 & 0.8425 & 0.7492 & 0.6971 & 0.9381 \\
& & & GPTQ & 0.8362 & 0.5329 & 0.6862 & 0.5068 & 0.7971 & 0.8520 & 0.7514 & 0.7089 & 0.9540 \\
& & & BlockRotation & 0.7839 & 0.5329 & 0.6772 & 0.5077 & 0.7921 & 0.8440 & 0.7421 & 0.6971 & 0.9381 \\
& & & ScaleSearch & 0.8317 & 0.5303 & 0.6802 & 0.5171 & 0.7837 & 0.8505 & 0.7459 & 0.7056 & 0.9496 \\
 & & & \soscalecell{SoScale}
& \soscalecell{0.8575} & \soscalecell{0.5334} & \soscalecell{0.6924}
& \soscalecell{0.5000} & \soscalecell{0.7875} & \soscalecell{0.8517}
& \soscalecell{0.7421} & \soscalecell{\underline{0.7092}} & \soscalecell{\underline{0.9545}} \\
 & & & \soscalecell{SoScale w/ Had.}
& \soscalecell{0.8271} & \soscalecell{0.5455} & \soscalecell{0.7010}
& \soscalecell{0.5137} & \soscalecell{0.8072} & \soscalecell{0.8636}
& \soscalecell{0.7584} & \soscalecell{\textbf{0.7166}} & \soscalecell{\textbf{0.9644}} \\
\cmidrule(lr){3-13}
& & \multirow{6}{*}{NVFP4}
& RTN & 0.8544 & 0.5542 & 0.7029 & 0.5401 & 0.8178 & 0.8560 & 0.7579 & 0.7262 & 0.9773 \\
& & & GPTQ & 0.8582 & 0.5515 & 0.7117 & 0.5247 & 0.8178 & 0.8584 & 0.7573 & 0.7257 & 0.9765 \\
& & & BlockRotation & 0.8423 & 0.5439 & 0.6966 & 0.5187 & 0.8093 & 0.8523 & 0.7600 & 0.7176 & 0.9657 \\
& & & ScaleSearch & 0.8620 & 0.5551 & 0.7100 & 0.5324 & 0.8215 & 0.8621 & 0.7535 & 0.7281 & 0.9798 \\
 & & & \soscalecell{SoScale}
& \soscalecell{0.8787} & \soscalecell{0.5556} & \soscalecell{0.7153}
& \soscalecell{0.5401} & \soscalecell{0.8349} & \soscalecell{0.8664}
& \soscalecell{0.7579} & \soscalecell{\textbf{0.7356}} & \soscalecell{\textbf{0.9899}} \\
 & & & \soscalecell{SoScale w/ Had.}
& \soscalecell{0.8582} & \soscalecell{0.5543} & \soscalecell{0.7204}
& \soscalecell{0.5384} & \soscalecell{0.8161} & \soscalecell{0.8651}
& \soscalecell{0.7628} & \soscalecell{\underline{0.7308}} & \soscalecell{\underline{0.9834}} \\
\cmidrule(lr){3-13}
& & \multirow{6}{*}{HiF4}
& RTN & 0.8529 & 0.5555 & 0.7053 & 0.5119 & 0.8056 & 0.8541 & 0.7622 & 0.7211 & 0.9704 \\
& & & GPTQ & 0.8643 & 0.5577 & 0.7109 & 0.5273 & 0.8173 & 0.8630 & 0.7617 & 0.7289 & 0.9809 \\
& & & BlockRotation & 0.8567 & 0.5475 & 0.7058 & 0.5154 & 0.8081 & 0.8587 & 0.7584 & 0.7215 & 0.9710 \\
& & & ScaleSearch & 0.8552 & 0.5587 & 0.7093 & 0.5213 & 0.8081 & 0.8554 & 0.7617 & 0.7242 & 0.9747 \\
 & & & \soscalecell{SoScale}
& \soscalecell{0.8696} & \soscalecell{0.5582} & \soscalecell{0.7154}
& \soscalecell{0.5299} & \soscalecell{0.8165} & \soscalecell{0.8590}
& \soscalecell{0.7579} & \soscalecell{\underline{0.7295}} & \soscalecell{\underline{0.9817}} \\
 & & & \soscalecell{SoScale w/ Had.}
& \soscalecell{0.8605} & \soscalecell{0.5563} & \soscalecell{0.7163}
& \soscalecell{0.5367} & \soscalecell{0.8161} & \soscalecell{0.8694}
& \soscalecell{0.7617} & \soscalecell{\textbf{0.7310}} & \soscalecell{\textbf{0.9838}} \\
\midrule
\multirow{19}{*}{\raisebox{-0.45\normalbaselineskip}{Qwen3-14B}}
& -- & BF16 & --
& 0.8863 & 0.6094 & 0.7710 & 0.5845 & 0.8418 & 0.8936 & 0.8020 & 0.7698 & 1.0000 \\
\cmidrule(lr){2-13}
& \multirow{18}{*}{W4A4}
& \multirow{6}{*}{MXFP4}
& RTN & 0.8408 & 0.5792 & 0.7243 & 0.5623 & 0.8186 & 0.8728 & 0.7671 & 0.7379 & 0.9585 \\
& & & GPTQ & 0.8711 & 0.5794 & 0.7313 & 0.5546 & 0.8157 & 0.8804 & 0.7753 & 0.7440 & 0.9664 \\
& & & BlockRotation & 0.8400 & 0.5759 & 0.7281 & 0.5469 & 0.8186 & 0.8798 & 0.7780 & 0.7382 & 0.9589 \\
& & & ScaleSearch & 0.8522 & 0.5839 & 0.7334 & 0.5631 & 0.8119 & 0.8789 & 0.7677 & 0.7416 & 0.9633 \\
 & & & \soscalecell{SoScale}
& \soscalecell{0.8726} & \soscalecell{0.5811} & \soscalecell{0.7321}
& \soscalecell{0.5691} & \soscalecell{0.8215} & \soscalecell{0.8865}
& \soscalecell{0.7726} & \soscalecell{\underline{0.7479}} & \soscalecell{\underline{0.9716}} \\
 & & & \soscalecell{SoScale w/ Had.}
& \soscalecell{0.8658} & \soscalecell{0.5898} & \soscalecell{0.7414}
& \soscalecell{0.5546} & \soscalecell{0.8203} & \soscalecell{0.8847}
& \soscalecell{0.7797} & \soscalecell{\textbf{0.7480}} & \soscalecell{\textbf{0.9717}} \\
\cmidrule(lr){3-13}
& & \multirow{6}{*}{NVFP4}
& RTN & 0.8757 & 0.5937 & 0.7496 & 0.5529 & 0.8224 & 0.8887 & 0.7791 & 0.7517 & 0.9765 \\
& & & GPTQ & 0.8719 & 0.5993 & 0.7554 & 0.5546 & 0.8338 & 0.8850 & 0.7922 & 0.7560 & 0.9821 \\
& & & BlockRotation & 0.8817 & 0.5847 & 0.7418 & 0.5478 & 0.8047 & 0.8826 & 0.7927 & 0.7480 & 0.9717 \\
& & & ScaleSearch & 0.8772 & 0.5976 & 0.7550 & 0.5683 & 0.8253 & 0.8856 & 0.7862 & 0.7565 & 0.9827 \\
 & & & \soscalecell{SoScale}
& \soscalecell{0.8901} & \soscalecell{0.5991} & \soscalecell{0.7580}
& \soscalecell{0.5811} & \soscalecell{0.8291} & \soscalecell{0.8890}
& \soscalecell{0.7851} & \soscalecell{\underline{0.7616}} & \soscalecell{\underline{0.9894}} \\
 & & & \soscalecell{SoScale w/ Had.}
& \soscalecell{0.8954} & \soscalecell{0.5956} & \soscalecell{0.7615}
& \soscalecell{0.5802} & \soscalecell{0.8338} & \soscalecell{0.8896}
& \soscalecell{0.7894} & \soscalecell{\textbf{0.7636}} & \soscalecell{\textbf{0.9920}} \\
\cmidrule(lr){3-13}
& & \multirow{6}{*}{HiF4}
& RTN & 0.8749 & 0.5987 & 0.7564 & 0.5657 & 0.8333 & 0.8936 & 0.7922 & 0.7593 & 0.9863 \\
& & & GPTQ & 0.8946 & 0.5993 & 0.7564 & 0.5700 & 0.8304 & 0.8896 & 0.7889 & 0.7613 & 0.9890 \\
& & & BlockRotation & 0.8795 & 0.5927 & 0.7512 & 0.5794 & 0.8367 & 0.8893 & 0.7905 & 0.7599 & 0.9871 \\
& & & ScaleSearch & 0.8772 & 0.6007 & 0.7558 & 0.5700 & 0.8375 & 0.8899 & 0.7927 & 0.7605 & 0.9880 \\
 & & & \soscalecell{SoScale}
& \soscalecell{0.8840} & \soscalecell{0.5965} & \soscalecell{0.7559}
& \soscalecell{0.5922} & \soscalecell{0.8422} & \soscalecell{0.8911}
& \soscalecell{0.7943} & \soscalecell{\textbf{0.7652}} & \soscalecell{\textbf{0.9940}} \\
 & & & \soscalecell{SoScale w/ Had.}
& \soscalecell{0.8954} & \soscalecell{0.6000} & \soscalecell{0.7595}
& \soscalecell{0.5700} & \soscalecell{0.8392} & \soscalecell{0.8890}
& \soscalecell{0.7884} & \soscalecell{\underline{0.7631}} & \soscalecell{\underline{0.9913}} \\
\bottomrule
\end{tabular}
}
\caption{
Downstream accuracy under W4A4 hierarchical FP4 quantization; \textbf{Avg.} averages tasks and \textbf{Rel.} normalizes by the BF16 average for the same model.
Highlighted rows are SoScale variants, \textbf{w/ Had.} denotes block Hadamard preprocessing, bold marks the best \textbf{Avg.} in each model-format group, and underlining marks the second best.
}
\label{tab:main-downstream-results}
\end{table*}

\begin{table*}[t]
\centering
\small
\renewcommand{\arraystretch}{0.89}
\setlength{\tabcolsep}{3pt}
\resizebox{\linewidth}{!}{
\begin{tabular}{l c l l ccccccc cc}
\toprule
\textbf{Model} & \textbf{Bit} & \textbf{Format} & \textbf{Method}
& \textbf{GSM8K} & \textbf{HellaSwag} & \textbf{MMLU}
& \textbf{ARC-C} & \textbf{ARC-E} & \textbf{BoolQ} & \textbf{PIQA}
& \textbf{Avg.} & \textbf{Rel.} \\
\midrule
\multirow{9}{*}{Qwen3-8B}
& \multirow{9}{*}{W4A4}
& \multirow{3}{*}{MXFP4}
& ScaleSearch-w & 0.8241 & 0.5273 & 0.6738 & 0.5009 & 0.7774 & 0.8492 & 0.7492 & 0.7003 & 0.9424 \\
& & & \soscalecell{SoScale-w}
& \soscalecell{0.8317} & \soscalecell{0.5292} & \soscalecell{0.6856}
& \soscalecell{0.5000} & \soscalecell{0.7997} & \soscalecell{0.8508}
& \soscalecell{0.7405} & \soscalecell{0.7054} & \soscalecell{0.9492} \\
& & & \soscalecell{SoScale-w w/ Had.}
& \soscalecell{0.8461} & \soscalecell{0.5438} & \soscalecell{0.6966}
& \soscalecell{0.5205} & \soscalecell{0.8064} & \soscalecell{0.8635}
& \soscalecell{0.7601} & \soscalecell{\textbf{0.7195}} & \soscalecell{\textbf{0.9683}} \\
\cmidrule(lr){3-13}
& & \multirow{3}{*}{NVFP4}
& ScaleSearch-w & 0.8605 & 0.5533 & 0.7124 & 0.5341 & 0.8114 & 0.8575 & 0.7655 & 0.7278 & 0.9795 \\
& & & \soscalecell{SoScale-w}
& \soscalecell{0.8726} & \soscalecell{0.5542} & \soscalecell{0.7082}
& \soscalecell{0.5384} & \soscalecell{0.8157} & \soscalecell{0.8602}
& \soscalecell{0.7459} & \soscalecell{0.7279} & \soscalecell{0.9796} \\
& & & \soscalecell{SoScale-w w/ Had.}
& \soscalecell{0.8560} & \soscalecell{0.5541} & \soscalecell{0.7121}
& \soscalecell{0.5410} & \soscalecell{0.8136} & \soscalecell{0.8657}
& \soscalecell{0.7579} & \soscalecell{\textbf{0.7286}} & \soscalecell{\textbf{0.9806}} \\
\cmidrule(lr){3-13}
& & \multirow{3}{*}{HiF4}
& ScaleSearch-w & 0.8400 & 0.5567 & 0.7074 & 0.5060 & 0.8035 & 0.8514 & 0.7650 & 0.7186 & 0.9670 \\
& & & \soscalecell{SoScale-w}
& \soscalecell{0.8613} & \soscalecell{0.5574} & \soscalecell{0.7124}
& \soscalecell{0.5230} & \soscalecell{0.8144} & \soscalecell{0.8636}
& \soscalecell{0.7557} & \soscalecell{0.7268} & \soscalecell{0.9781} \\
& & & \soscalecell{SoScale-w w/ Had.}
& \soscalecell{0.8582} & \soscalecell{0.5591} & \soscalecell{0.7140}
& \soscalecell{0.5444} & \soscalecell{0.8186} & \soscalecell{0.8700}
& \soscalecell{0.7655} & \soscalecell{\textbf{0.7328}} & \soscalecell{\textbf{0.9861}} \\
\midrule
\multirow{9}{*}{Qwen3-14B}
& \multirow{9}{*}{W4A4}
& \multirow{3}{*}{MXFP4}
& ScaleSearch-w & 0.8567 & 0.5795 & 0.7328 & 0.5469 & 0.8140 & 0.8670 & 0.7584 & 0.7365 & 0.9567 \\
& & & \soscalecell{SoScale-w}
& \soscalecell{0.8749} & \soscalecell{0.5762} & \soscalecell{0.7318}
& \soscalecell{0.5700} & \soscalecell{0.8249} & \soscalecell{0.8813}
& \soscalecell{0.7715} & \soscalecell{0.7472} & \soscalecell{0.9707} \\
& & & \soscalecell{SoScale-w w/ Had.}
& \soscalecell{0.8673} & \soscalecell{0.5877} & \soscalecell{0.7421}
& \soscalecell{0.5555} & \soscalecell{0.8236} & \soscalecell{0.8853}
& \soscalecell{0.7829} & \soscalecell{\textbf{0.7492}} & \soscalecell{\textbf{0.9732}} \\
\cmidrule(lr){3-13}
& & \multirow{3}{*}{NVFP4}
& ScaleSearch-w & 0.8901 & 0.5970 & 0.7569 & 0.5631 & 0.8237 & 0.8829 & 0.7753 & 0.7556 & 0.9815 \\
& & & \soscalecell{SoScale-w}
& \soscalecell{0.8878} & \soscalecell{0.5971} & \soscalecell{0.7584}
& \soscalecell{0.5794} & \soscalecell{0.8249} & \soscalecell{0.8902}
& \soscalecell{0.7829} & \soscalecell{0.7601} & \soscalecell{0.9874} \\
& & & \soscalecell{SoScale-w w/ Had.}
& \soscalecell{0.8878} & \soscalecell{0.5975} & \soscalecell{0.7587}
& \soscalecell{0.5768} & \soscalecell{0.8333} & \soscalecell{0.8911}
& \soscalecell{0.7813} & \soscalecell{\textbf{0.7609}} & \soscalecell{\textbf{0.9885}} \\
\cmidrule(lr){3-13}
& & \multirow{3}{*}{HiF4}
& ScaleSearch-w & 0.8825 & 0.6002 & 0.7528 & 0.5717 & 0.8316 & 0.8838 & 0.7835 & 0.7580 & 0.9847 \\
& & & \soscalecell{SoScale-w}
& \soscalecell{0.8893} & \soscalecell{0.5954} & \soscalecell{0.7578}
& \soscalecell{0.5657} & \soscalecell{0.8413} & \soscalecell{0.8917}
& \soscalecell{0.7905} & \soscalecell{0.7617} & \soscalecell{0.9894} \\
& & & \soscalecell{SoScale-w w/ Had.}
& \soscalecell{0.8946} & \soscalecell{0.6008} & \soscalecell{0.7609}
& \soscalecell{0.5836} & \soscalecell{0.8380} & \soscalecell{0.8902}
& \soscalecell{0.7954} & \soscalecell{\textbf{0.7662}} & \soscalecell{\textbf{0.9953}} \\
\bottomrule
\end{tabular}
}
\caption{
W4A4 comparison with direct RTN activation quantization.
The suffix ``-w'' denotes variants that use weight-side scale search while keeping activations on a direct RTN path; \textbf{w/ Had.} denotes block Hadamard preprocessing before weight quantization.
Highlighted cells denote SoScale variants, and bold values mark the best average accuracy within each model-format group.
}
\label{tab:weight-search-only-results}
\end{table*}

\begin{table}[t]
\centering
\small
\renewcommand{\arraystretch}{0.89}
\setlength{\tabcolsep}{5pt}
\resizebox{\columnwidth}{!}{
\begin{tabular}{l l c c c c}
\toprule
\textbf{Format} & \textbf{Variant} & \textbf{Loss} & \textbf{Rec.} & \textbf{Avg.} & \textbf{Rel.} \\
\midrule
\multirow{4}{*}{MXFP4}
& RTN & N/A & No & 0.6971 & 0.9381 \\
& ScaleSearch & MSE & No & 0.7056 & 0.9496 \\
& MSE + recovery & MSE & \checkmark & 0.7061 & 0.9503 \\
% & \soscalecell{SoScale w/o rec.} & \soscalecell{Hessian} & \soscalecell{No} & \soscalecell{TBD} & \soscalecell{TBD} \\
& \soscalecell{SoScale} & \soscalecell{Hessian} & \soscalecell{\checkmark} & \soscalecell{0.7092} & \soscalecell{0.9545} \\
\cmidrule(lr){1-6}
\multirow{4}{*}{NVFP4}
& RTN & N/A & No & 0.7262 & 0.9773 \\
& ScaleSearch & MSE & No & 0.7281 & 0.9798 \\
& MSE + recovery & MSE & \checkmark & 0.7270 & 0.9784 \\
% & \soscalecell{SoScale w/o rec.} & \soscalecell{Hessian} & \soscalecell{No} & \soscalecell{TBD} & \soscalecell{TBD} \\
& \soscalecell{SoScale} & \soscalecell{Hessian} & \soscalecell{\checkmark} & \soscalecell{0.7356} & \soscalecell{0.9899} \\
\cmidrule(lr){1-6}
\multirow{4}{*}{HiF4}
& RTN & N/A & No & 0.7211 & 0.9704 \\
& ScaleSearch & MSE & No & 0.7242 & 0.9747 \\
& MSE + recovery & MSE & \checkmark & 0.7263 & 0.9774 \\
% & \soscalecell{SoScale w/o rec.} & \soscalecell{Hessian} & \soscalecell{No} & \soscalecell{TBD} & \soscalecell{TBD} \\
& \soscalecell{SoScale} & \soscalecell{Hessian} & \soscalecell{\checkmark} & \soscalecell{0.7295} & \soscalecell{0.9817} \\
\bottomrule
\end{tabular}
}
\caption{
Qwen3-8B ablation under W4A4 quantization; \textbf{Rec.} denotes the 2.2 error-propagation stage in Figure~\ref{fig:dynamic-scale-search-gptq}.
}
\label{tab:qwen3-8b-core-ablation}
\end{table}

\paragraph{SoScale improves downstream accuracy.}
Table~\ref{tab:main-downstream-results} reports downstream accuracy in the main W4A4 setting. Across both Qwen3 models and all three FP4 formats, a SoScale variant obtains the best average score in every model-format group, preserving 96.44\%--97.17\% of the BF16 average accuracy under MXFP4, 98.99\%--99.20\% under NVFP4, and 98.38\%--99.40\% under HiF4. Compared with the strongest non-SoScale baseline in each group, the best SoScale variant improves the average score by 0.21 to 0.77 percentage points; relative to local ScaleSearch, the gains are 0.47 to 1.10 points. The largest gains appear under MXFP4 because its E8M0 scale codes allow only coarse power-of-two adjustments: outliers can force a coarse FP4 grid that leaves dense values under-resolved, creating more room for Hessian-aware scale selection. NVFP4 and HiF4 provide stronger baselines and leave less headroom, yet SoScale still improves them, indicating that layer-level sensitivity remains useful even when the scale space is finer.

\paragraph{Hadamard gains are format-dependent.}
BlockRotation alone gives mixed improvements: it can help in some model-format groups, but it does not consistently outperform ScaleSearch and sometimes trails RTN. Combining Hadamard preprocessing with SoScale is most useful for MXFP4, where it gives the best average score for both Qwen3-8B and Qwen3-14B, and it also gives the best NVFP4 result on Qwen3-14B. However, the best row is not always the Hadamard variant: Qwen3-8B under NVFP4 and Qwen3-14B under HiF4 are best with SoScale alone. This format dependence is consistent with recent microscaling FP4 analysis, which highlights MXFP4's power-of-two scale error and NVFP4's distinct small-group behavior~\citep{egiazarian2026bridginggappromiseperformance}. Overall, preprocessing changes the distribution presented to the quantizer, whereas SoScale selects the hardware-valid scale codes used in the final FP4 representation. This explains why rotations can help MXFP4, but do not replace Hessian-aware scale selection.

\paragraph{Direct activation RTN remains viable.}
Table~\ref{tab:weight-search-only-results} evaluates a deployment-oriented variant in which activations use direct RTN quantization instead of activation-side scale search. This setting keeps the activation path aligned with hardware-supported FP4 formats and avoids introducing format-specific activation search into the runtime. Under this direct activation path, the best SoScale-w variant still outperforms ScaleSearch-w in every model-format group, with gains ranging from 0.08 to 1.92 percentage points in average accuracy. Hadamard preprocessing remains useful here because the paired orthogonal transforms applied to weights and activations preserve the linear map while spreading outliers on both sides of the multiplication. This is especially helpful when activations follow a direct RTN path, where no activation-side search is available to correct outlier-induced scale choices; the clearest gains again appear for MXFP4. Some direct-RTN rows also exceed their counterparts in Table~\ref{tab:main-downstream-results}. This does not imply that direct activation RTN is uniformly better; rather, the online activation rule is a local heuristic and is not jointly optimized with the weight scale codes or downstream loss. When activations are already well matched to the target FP4 format, avoiding this extra activation-side adaptation can be beneficial, while SoScale continues to address the dominant weight-side scale-code error.

\paragraph{Ablating search and recovery.}
Table~\ref{tab:qwen3-8b-core-ablation} compares MSE scale search, MSE search with blockwise recovery, and full SoScale on Qwen3-8B under fixed FP4 formats. Adding recovery to the MSE objective gives mixed gains over ScaleSearch: it is marginal under MXFP4, decreases NVFP4, and improves HiF4. In contrast, full SoScale is best for all three formats, indicating that recovery is most effective when coupled with Hessian-aware scale-code selection rather than a local MSE objective.

\section{Conclusion}

We presented \textbf{SoScale}, a second-order scale-code search method for hierarchical FP4 quantization. SoScale addresses a practical failure mode of low-bit floating-point formats: range-preserving or locally MSE-optimal scales can place the FP4 grid poorly for layer outputs, especially when outliers interact with discrete scale codes. By optimizing hardware-valid weight scale codes under a Hessian-aware layer reconstruction objective and interleaving this search with GPTQ-style error recovery, SoScale improves W4A4 quantization without changing the target FP4 representation. Experiments on Qwen3-8B and Qwen3-14B across MXFP4, NVFP4, and HiF4 show consistent gains over RTN, GPTQ, local ScaleSearch, and BlockRotation baselines, with the largest headroom under MXFP4 and persistent gains for stronger formats. The direct-activation setting further shows that SoScale can improve weight quantization while keeping the activation path simple and hardware aligned. These results suggest that scale-code selection should be treated as a layer-aware optimization problem, rather than merely a local tensor-fitting rule, for accurate low-bit floating-point LLM inference.

\clearpage
\bibliography{./iclr2025_conference}

% If so, you can uncomment the following line and adjust the path to include it.

\end{document}